\documentclass[11pt]{article}

\usepackage{a4wide,graphicx,times,amsmath} 
\usepackage{algorithm,algorithmic}
\usepackage{multirow}

\usepackage{subfigure}
\usepackage{pstricks}
\usepackage{pstricks-add}
\usepackage{pst-node}
\usepackage{pst-plot}
\usepackage{listings}
\usepackage{color}

\newcommand{\pmin}{p_{\min}}
\newcommand{\pmax}{p_{\max}}

\newrgbcolor{lightblue}{0.60 0.60 1.00}
\newrgbcolor{lightred}{1.00 0.60 .60}

\begin{document}

\title{A Protocol for Self-Synchronized Duty-Cycling in Sensor Networks: Generic Implementation in Wiselib}

\author{H.~Hern{\'a}ndez$^1$, T.~Baumgartner$^2$, M.~J.~Blesa$^1$, C.~Blum$^1$, A.~Kr{\"o}ller$^2$ and S.~P.~Fekete$^2$ \\
~\\
$^1$ ALBCOM Research Group\\ 
Universitat Polit{\`e}cnica de Catalunya, Barcelona, Spain \\
{\sf \{hherandez,mjblesa,cblum\}@lsi.upc.edu} \\
~\\
$^2$ Algorithms Group\\
Braunschweig Institute of Technology, Braunschweig, Germany \\
{\sf \{t.baumgartner,a.kroeller,s.fekete\}@tu-bs.de}}

\date{}

\maketitle

\begin{abstract}
In this work we present a protocol for self-synchronized duty-cycling in wireless sensor networks with energy harvesting capabilities. The protocol is implemented in Wiselib, a library of generic algorithms for sensor networks. Simulations are conducted with the sensor network simulator Shawn. They are based on the specifications of real hardware known as iSense sensor nodes. The experimental results show that the proposed mechanism is able to adapt to changing energy availabilities. Moreover, it is shown that the system is very robust against packet loss.
\end{abstract}

\section{Introduction}

Sensor networks~\cite{WagWat07:book} consist of a set of small, autonomous devices which may be used, for example, for environmental monitoring, patient monitoring in health care, and industrial machinery surveillance. As sensor nodes may be distributed in areas without power supply, or they may be mobile, they are generally equipped with batteries, which makes energy a scarce resource. A basic idea for saving energy is to periodically switch off the sensor nodes. The mechanism that establishes the alternation between the active and inactive states is generally called \emph{duty-cycling} (see, for example,~\cite{TiaGeo03:ds}). In some cases, duty-cycling is energy-aware (see, for example,~\cite{HsuEtAl06:islped,raghunathan2005design,KanEtAl07:powman}). The main disadvantage of these approaches is that they require a quite regular pattern for the availability of energy from the environment. 

\begin{figure}[!t]
    \centering
        \includegraphics[width=5.5cm,angle=-90]{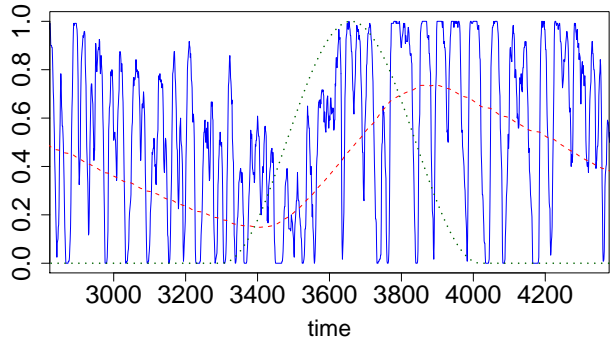}
    \caption{A first glimpse of the working of energy-aware, self-synchronized duty-cycling.\label{fig:generic-behavior}}
\end{figure}

In previous work~\cite{HugBlu09:self,HugBlu09:static,HugBlu09:async} we introduced and studied a possible technique for energy-aware duty-cycling in (mobile) sensor networks with energy harvesting capabilities. This system is inspired by self-synchronized sleeping patterns of natural ant colonies~\cite{FraBra87:rhythmical}. The focus of these first studies was purely on the swarm intelligence aspects of the proposed system. The experiments were performed without considering, for example, packet loss, collisions and network failures. Before we outline the contributions of this work, we introduce already a glimpse of the basic behavior of this previously introduced system; see Figure~\ref{fig:generic-behavior}. The solid line shows the fraction of active nodes over time, whereas the slashed line shows the average battery level of the nodes over time. Finally, the dotted line represents the sun power that is used to establish the amount of energy which can be harvested by the sensor nodes at each time step. Note that all three measures are scaled to $[0,1]$. At this point we would like the interested reader to understand the following two aspects. First, self-synchronized duty-cycling is indicated by the repetitive appearance of activity peaks over time (see solid line). Second, the adaptation to changing energy conditions is indicated by the changing height of the activity peaks. At times of lower battery levels, activity peaks are lower as well. This is the mechanism used by the sensor nodes to adapt to varying energy conditions.

\noindent {\bf Contribution of this Work.} The first contribution of this work is the design of a protocol that captures the essential aspects of the swarm intelligence system previously proposed in~\cite{HugBlu09:self,HugBlu09:static,HugBlu09:async}. The second contribution consists in the implementation of this protocol in Wiselib~\cite{wiselib}, a library of generic algorithms for wireless sensor networks. Finally, we experimentally test our duty-cycling protocol in a real scenario, simulating iSense sensor nodes from Coalesenses GmbH~\cite{coalesenses} with the network simulator Shawn~\cite{shawn}.

The organization of the paper is as follows. In Section~\ref{sec:wiselib} the extension of Wiselib for duty-cycling algorithms is introduced, followed by the description of the protocol for self-synchronized duty-cycling in Section~\ref{sec:concept}. Finally, experimental results are presented in Section~\ref{sec:simulations}, and conclusions and an outlook are given in Section~\ref{sec:conclusions}. 

\section{Wiselib: Duty-Cycling Concept}
    \label{sec:wiselib}

The Wiselib~\cite{wiselib,wiselib-web} is a generic algorithm library for heterogeneous wireless sensor networks. The main objective concerns the implementation of algorithms that may be applied on different hardware and software platforms. Not only real sensor hardware such as MicaZ or TelosB nodes are supported by Wiselib, but also simulation environments such as Shawn and TOSSIM. The algorithms included in the Wiselib are organized in topics according to their functionality. In order to abstract the algorithms from the hardware and the operating system, a set of connectors specifies interfaces for interacting with them. A connector is also defined to interact with wireless sensor network simulators such as Shawn~\cite{shawn}. Those connectors are defined such that the same algorithm can be run on a physical platform or on a simulator. 

The design of a new algorithm class in Wiselib requires the definition of a concept, that is, a specification of how an algorithm looks and behaves. In this paper we provide a generic concept for energy-aware duty-cycling algorithms, and provide one model for this concept (see Section \ref{sec:concept}). In particular, a duty-cycling algorithm must assist sensor nodes in their decision of being active or inactive. This is handled via a call-back to the sensor node when it is supposed to change mode. Based on this call-back, the sensor node is then responsible for the executing the corresponding action.

Therefore, a duty-cycling algorithm has basically two functionalities: It can be enabled and disabled, and a call-back can be (un)registered to indicate changes in activity. The concept looks as follows:
\begin{lstlisting}[language=C++,basicstyle=\scriptsize]
concept DutyCycling {
  enum DutyCyclingActivity {
    DC_ACTIVE, DC_INACTIVE
  };
  void enable(void);
  void disable(void);
  template<class Callee, void (Callee::*TMethod)(int)>
    int reg_changed_callback( Callee *obj_pnt )
  void unreg_recv_callback(int);
};
\end{lstlisting}

With the aid of this generic concept, it is possible to cover a broad range of duty-cycling algorithms. The exact behavior of a potential duty-cycling model is not mandatory. It can be asynchronous or synchronized, it may rely on exact time-synchronization or do not have any requirements. The important aspect is that the method signatures from the concept are implemented, so that it can be passed to other algorithms as a template parameter.

\section{Proposed Duty-Cycling Model}
    \label{sec:concept}

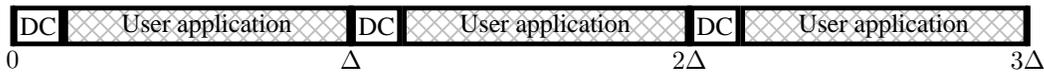
\begin{figure*}[t]
\centering
\scalebox{0.9}{
\psset{xunit=5cm,yunit=2cm,runit=5cm}
\begin{pspicture}(-0.1,-0.1)(3.2,0.5)
\psgrid[gridcolor=white,subgridcolor=white,subgriddots=10,gridwidth=1.5pt,subgridwidth=1.5pt,gridlabels=0pt](0,0)(3,0.3)
\psframe[linewidth=2pt,fillstyle=none,fillcolor= lightgray,linecolor=black](0,0)(0.15,0.3)
\psframe[linewidth=2pt,fillstyle=crosshatch,fillcolor= lightgray,linecolor=black,hatchcolor=lightgray](0.15,0)(1,0.3)
\psline[linewidth=2pt,linecolor=black](0.0,0)(0.0,0.3)

\psframe[linewidth=2pt,fillstyle=none,fillcolor= lightgray,linecolor=black](1,0)(1.15,0.3)
\psframe[linewidth=2pt,fillstyle=crosshatch,fillcolor= lightgray,linecolor=black,hatchcolor=lightgray](1.15,0)(2,0.3)
\psline[linewidth=2pt,linecolor=black](1.0,0)(1.0,0.3)

\psframe[linewidth=2pt,fillstyle=none,fillcolor= lightgray,linecolor=black](2,0)(2.15,0.3)
\psframe[linewidth=2pt,fillstyle=crosshatch,fillcolor= lightgray,linecolor=black,hatchcolor=lightgray](2.15,0)(3,0.3)
\psline[linewidth=2pt,linecolor=black](2.0,0)(2.0,0.3)

\psline[linewidth=2pt,linecolor=black](3.0,0)(3.0,0.3)

\rput(0,-0.1){$0$}
\rput(1,-0.1){$\Delta$}
\rput(2,-0.1){$2\Delta$}
\rput(3,-0.1){$3\Delta$}
\rput(0.075,0.15){DC}
\rput(0.575,0.15){User application}
\rput(1.075,0.15){DC}
\rput(1.575,0.15){User application}
\rput(2.075,0.15){DC}
\rput(2.575,0.15){User application}
\end{pspicture}}
\caption{Division of time between the duty-cycling mechanism and user applications. The protocol works in a sequence of time periods of length $\Delta$. In each period, the first phase is dedicated to duty-cycling (DC), and the second phase to the user application.\label{fig:cycle-scheduling}}
\end{figure*}

As mentioned before, in~\cite{HugBlu09:self,HugBlu09:static,HugBlu09:async} a swarm intelligence technique with the potential application of self-synchronized duty-cycling in (mobile) wireless sensor networks with energy harvesting capabilities was introduced. This section aims at designing and implementing a duty-cycling protocol on the basis of this work. The current version of the protocol assumes that there is a time synchronization algorithm executed by a lower layer of the network. The protocol works in periods. Each period has a length of $\Delta$ time units (say, seconds). Each period is divided in two phases: the first phase is dedicated to actions concerning the management of duty-cycling, whereas the second phase is dedicated to application-specific tasks that sensor nodes must perform (see Figure~\ref{fig:cycle-scheduling}). The first phase of each period is very short. In this phase all nodes may receive transmissions from neighboring nodes and themselves they execute one duty-cycling event. The outcome of the first phase decides if a node will be \emph{active} or \emph{inactive} for the rest of the corresponding period. In case of being active a node is available for user-defined applications (environmental data monitoring, tracking, etc). However, if the state of a sensor node is set to inactive the node will turn off the radio and will sleep until the start of the following period.

In the following we focus on the description of the duty-cycling algorithm executed in the first phase of each period. This algorithm consists in a so-called \emph{duty-cycling event} that is executed by each sensor node $i$ exactly once. The time of executing this event is, at the moment, randomly chosen by each sensor node within the first phase of each period. Each sensor node $i$ maintains a real-valued state variable $S_i$. The value of this state variable is initially set to the so-called \emph{activity threshold} $S_{\text{\scriptsize act}}$, which is a parameter of the mechanism. Moreover, $a_i$ is a binary variable whose value determines if the sensor node $i$ is \emph{active} ($a_i = 1$) or \emph{inactive} ($a_i = 0$) in phase of two of the corresponding period. The value of the variable $a_i$ is determined as follows:
\begin{equation}\label{eqn:calculate-activity}
  a_i :=  \Phi(S_i - \theta_{\text{\scriptsize act}}) \enspace, 
\end{equation}
where $\theta_{\text{\scriptsize act}}$ is the so-called activation threshold, and $\Phi(x) = 1$ if $x \geq 0$, and $\Phi(x) = 0$ otherwise. Note that an inactive sensor node can return to the active state either due to local interactions (as explained below in Eq.~\ref{eqn:state-update}) or spontaneously with a probability $p_a$ and an activity level $S_a$.

In addition, each sensor node maintains a queue $Q_i$ for incoming duty-cycling messages from neighboring sensor nodes. Each message $m\in Q_i$ contains a single field $m_{\text{\scriptsize activity}}$, which is set to the activity $S_j$ of the sensor node $j$ that has sent the message. When sensor node $i$ executes its duty-cycling event, the value of state variable $S_i$ is updated depending on the messages in $Q_i$. Subsequently sensor node $i$ sends a duty-cycling message, containing the new value of $S_i$, using a certain transmission power level. Note that the choice of the transmission power level is a crucial component for the working of our duty-cycling technique. More specifically, the value of state variable $S_i$ of a sensor node $i$ is computed as follows:
\begin{equation}\label{eqn:state-update}
 S_i := tanh(g\cdot (S_i + \sum_{m\in Q_i}m_{\text{\scriptsize activity}})) \enspace,
\end{equation}
where $g$ is a gain parameter whose value determines how fast the value of variable $S_i$ diminishes. After this update, all messages are deleted from $Q_i$, that is $Q_i := \emptyset$. Note that with this update the value $S_i$ of an inactive sensor may increase sufficiently enough in order to be greater than the activity threshold $S_{\text{\scriptsize act}}$. 

For the working of the duty-cycling mechanism, the choice of the power level for the transmission of the duty-cycling messages plays a crucial role. Here we assume a standard antenna model which allows sensor nodes---for each transmission---to choose from a finite set $P=\{P^1,\ldots,P^n\}$ of different transmission power levels.\footnote{Popular sensor hardware such as iSense nodes or SunSPOTs are equipped with such antennas.} More specifically, the choice of a sensor node $i$ depends on its battery level, which is denoted by $b_i \in [0,1]$. Hereby, $b_i = 1$ corresponds to a full battery. In the following we first outline the determination of a so-called \emph{ideal transmission power level}, which then leads to the choice of the real transmission power level. The ideal transmission power level ($p_i$) of a sensor $i$ depends on the current battery level: $p_i := \pmin \cdot (1 - b_i) + \pmax \cdot b_i$, where $\pmin$, respectively $\pmax$, are parameters that fix the minimum, respectively maximum, transmission power levels. Only when batteries are fully charged the ideal transmission power level may reach $\pmax$. The ideal transmission power level is then translated into the \emph{real transmission power level} $(T_i)$ as follows:  $T_i :=  P^k \in P$ such that
\begin{equation}\label{eqn:power-level}
 p_i \in \left( (P^{k-1} + P^{k})/2, (P^{k} + P^{k+1})/2 \right]
\end{equation}
At this point it is important to realize that the transmission power level $T_i$ is used only for sending the duty-cycling message. For other messages during the second phase of each period, the user application is responsible for choosing transmission power levels. The duty-cycling event described above is summarized in Algorithm~\ref{algo:basic}. As mentioned above, the battery level of the sensor nodes is responsible for their choice of a transmission power level for sending the duty-cycling message. Therefore, the battery level of course affects the communication topology in the context of the duty-cycling mechanism. 

\begin{algorithm}[t]
\caption{Duty-cycling event of a sensor node $i$ \label{algo:basic}}
\begin{algorithmic}[1]
\STATE Calculate $a_i$ (see Eq.~\ref{eqn:calculate-activity})
\IF{$a_i = 0$}
  \STATE Draw a random number $p \in [0,1]$
  \STATE {\bf if} $p \leq p_a$ {\bf then} $S_i := S_a$ and $a_i := 1$ {\bf endif}
\ENDIF
\STATE Determine transmission power level $T_i$ (see Eq.~\ref{eqn:power-level})
\STATE Compute new value for state variable $S_i$ (see Eq.~\ref{eqn:state-update})
\STATE Send duty cycling message $m$ with $m_{\text{\scriptsize activity}}:=S_i$ with transmission power level $T_i$\end{algorithmic}
\end{algorithm}

\section{Experimental Evaluation}
    \label{sec:simulations}

In the following we first describe the experimental setup and the experimentation environment before we present the obtained results. The implementation of the presented protocol in the Wiselib provides us with options for executing it on real test-beds but also to perform simulations with some sensor network simulators. In the context of this paper we decided for the second option. More specifically we used the sensor network simulator Shawn~\cite{shawn}, which is a discrete event simulator with a very high level of parameterization which is able to execute algorithms from the Wiselib. The user can easily run experiments simulating the behavior of different sensor nodes and also add own sensor node specifications. A peculiarity of Shawn is the fact that packet collisions are not explicitly considered. Instead Shawn simulates these collisions and the consequent packet loss under different constraints and in different scenarios. Thus, any packet-loss model can be implemented.

We decided to experiment with \emph{iSense} sensor node hardware from Coalesenses GmbH~\cite{coalesenses}. For this purpose we added the specification of \emph{iSense} nodes to Shawn. These sensor nodes use a Jennic JN5139 chip, a solution that combines the controller and the wireless communication transceiver in a single chip. The controller has a 32-bit RISC architecture and runs at 16Mhz. It is equipped with 96kb of RAM and 128kb of serial flash. The maximum transmission power level of iSense nodes reaches a distance of about 500m in all directions in open air conditions. Note that iSense nodes are being used by two of the currently largest European projects on sensor networks: WISEBED~\cite{wisebed, wisebed-project} and FRONTS~\cite{fronts-project}. In our simulations, iSense nodes are supposed to be equipped with solar panels. According to their documentation, iSense nodes require $0.025mA$ to work without using any additional peripheral such as the radio or the sensing devices. The state in which the radio is also turned on requires a power supply of $12.8mA$. Additionally, to receive or send a message with $4$ bytes of information, as required by duty-cycling messages, implies a consumption of $7.43\mu C$. The batteries have a maximum capacity of $2600\mu C$. Energy harvesting by solar panels can reach a maximum nominal value of $1.6\text{W}$. This information is summarized in Table~\ref{tab:energy-consumption}. Finally, let us mention that iSense nodes offer $6$ possible transmission power levels, in addition to the state in which the radio is turned off. The five transmission power levels other than the maximum one are obtained by reducing the maximum transmission power level by $\frac{1}{6}$, $\frac{2}{6}$, $\frac{3}{6}$, $\frac{4}{6}$, and $\frac{5}{6}$.

\begin{table}[!t]
\caption{Power devices and parameters for the energy model.\label{tab:energy-consumption}}
\centering
\scalebox{0.8}{
\begin{tabular}{ccc}
\hline
Data&$\;\;$& Device specifications\\
\hline
Tx/Rx (4 bytes) && $7.43\mu\text{C}$\\
Radio On && $12.8\text{mA}$\\
Radio Off && $0.025\text{mA}$\\
Battery capacity&&$2600\mu\text{C}$\\
Energy harvesting (f)&&$1.6\text{W}$\\
Max.~Tx Power&&$500\text{m}$\\
\hline
\end{tabular}}
\end{table}

One of the aspects that has not been described so far is the simulation of the light source for energy harvesting. This was done as follows. The light source at time $z > 0$ has an intensity of $s(z) \in [0,1]$. Hereby, $s(z) = 0$ corresponds to absolute darkness. In~\cite{HugBlu09:self} we described a model for the evolution of the sun light intensities, that is, for the evolution of $s(z)$ over time. Here we consider exactly the same model. Additionally, we assume a variable cloud density $c(z) \in [0,1]$. Depending on the technical characteristics of the solar panels used, a sensor node $i$ can transform a fraction $f$ of the available light intensity into energy:
\begin{equation}
\label{eq:harvesting}
e^{\text{\scriptsize harv}}_i((t-\Delta,t]) := f \cdot \int_{t-\Delta}^{t} s(z)\cdot (1-c(z)) dz
\end{equation}
In the experiments presented in this article we do not consider any specific user application, that is, the energy consumption of phase two of the proposed protocol must be simulated. This is done by removing an amount of $e_{\text{\scriptsize app}}$ of energy from the battery for each execution of phase two. The parameter values used for simulation are as follows:
\begin{center}
\scalebox{0.7}{
\begin{tabular}{|cccccc|}
\hline
$\pmin$ & $\pmax$ & $g$ & $p_{\text{\scriptsize a}}$&$e_{\text{\scriptsize app}}$&$f$ \\
\hline
$0.07$ & $0.14$ & $0.1$ & $0.001$ & $0.001$&$0.0027$\\
\hline
\end{tabular}}
\end{center}
It is important to note that the information which refers to the power profile of the iSense nodes is obtained by properly rescaling the values from Table~\ref{tab:energy-consumption} to the $[0,1]$ range that is used by the description of duty-cycling given in Section~\ref{sec:concept}.

\subsection{Experiments}

Assuming that $\Delta$---that is, the length of one period---corresponds to 60 seconds, the simulations that we conducted span $30$ days (each day consists of $1440$ periods). The first phase of each period, which is reserved for the duty-cycling events, was given $0.05$ seconds. Information about the state of the sensor nodes (active versus inactive) is collected at the start of each period. The most important measure taken is the \emph{mean activity} of the sensor network, which is measured---at any time---as follows:
\begin{equation}\label{eqn:at}
 A := \frac{1}{k} \sum_{i=1}^{k}a_i \in [0,1]
\end{equation}
Note that, the greater $A$ the more sensors are active at the specific time at which $A$ was determined. Self-synchronization behavior is characterized by an oscillating value of $A$ over time. This was shown already for a mobile sensor network with $k=120$ sensor nodes in Figure~\ref{fig:generic-behavior} of the introduction (see solid line). However, the results from that figure were obtained in a \emph{perfect environment} with no collisions or transmission failures and no propagation times. Moreover, the energy model that was used had no relation to real sensor node hardware. 

\begin{figure}[!t]
    \centering
      \includegraphics[width=5.5cm,angle=-90]{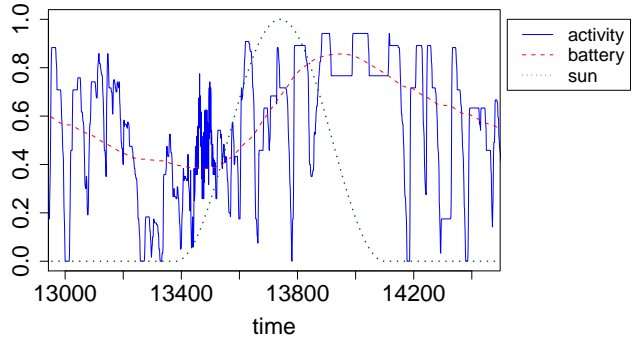}
    \caption{Simulation results (network size: $120$; 11th day of simulation). Solid line: evolution of the fraction of active nodes. Dashed line: average battery level. Dotted line: evolution of the sun light intensity.\label{fig:behavior}}
\end{figure}

The experiments that we present in this section aim at proving the applicability of the proposed mechanism in real sensor networks. All experiments are done on the basis of a static network of $k=120$ iSense nodes as simulated by the Shawn sensor network simulator. For the first experiment that we conducted we assumed a zero probability for packet collisions. Moreover, we assume a cloud density of zero, that is, $c(z) = 0$ for $z \geq 0$. Figure~\ref{fig:behavior} shows the obtained duty-cycling behavior. Again, the solid line shows the fraction of active sensor nodes over time, whereas the slashed line shows the average battery level of the sensor nodes over time. Finally, the dotted line represents the sun power that is used to establish the amount of energy which can be harvested by the sensor nodes at each time step. The graphic shows the behavior for one day of simulation, that is, 1440 periods. Self-synchronized duty-cycling is indicated by the appearance of activity peaks over time. It is remarkable how the system adapts to the available energy resources, reducing the height of the peaks when the battery level of the nodes is reduced. Note that when a lot of energy is available the system can even prescind from switching off sensor nodes. This can be seen by the existence of a large activity peak of about 200 periods of length located around period 14000. Note that for this experiment the average fraction of nodes that are active at each period is approximately $0.6$. This measure will henceforth be called the \emph{mean system activity}.

Concerning the energy spent by the duty-cycling protocol with respect to the user application, we took measures over the whole simulation of $43200$ periods (that is, 30 days).
\begin{center}
\scalebox{0.7}{
\begin{tabular}{|c|cccc|c|} \hline
       & \multicolumn{4}{c|}{\bf Duty-cycling} & {\bf User Appl.} \\ 
       & Tx & Rx & Idle & Active &  \\ \hline
Energy ($\%$) & 0.757 & 18.591 & 0.001 & 0.035 & 80.616 \\ \hline
\end{tabular}}
\end{center}
The energy spent by duty-cycling is hereby split into the "Idle" and "Active" states as well as the energy spent for transmitting the duty-cycling messages (Tx) and receiving duty-cycling messages (Rx). Note that message reception is the task which consumes most of the energy. In total, the duty-cycling mechanism consumes approximately $20\%$ of the total amount of spent energy. This may seem quite high at first. However, consider that this percentage strongly depends on the value of $e_{\text{\scriptsize app}}$, which we have set to a very moderate value of 0.001. Increasing this value will obviously cause the decrease of the percentage of energy spent by duty-cycling. 

After these initial studies we will now test the duty-cycling mechanism in two adversarial scenarios. In first place, it is shown how the system responds to situations with communication failures. In second place, the behavior of the system is studied in scenarios where energy harvesting is limited, for example, due to cloudy weather. Finally, we present a mechanism for the automatic parameter adaptation of the system for what concerns different network sizes. 

\subsubsection{Effect of Packet Loss}

With the next experiment we aim at studying the robustness of the system with respect to communication failures. The experiment consists in simulating the duty-cycling protocol under different packet loss rates. A packet loss rate of $p_{\text loss} \in [0,1]$ means that the probability of correctly receiving a message is $1 - p_{\text loss}$. We repeated the initial experiment as outlined above for all packet loss rates between 0 and 1, in steps of 0.01. The results are shown in terms of the obtained \emph{mean system activity} for each considered packet loss rate in Figure~\ref{fig:duty-experiments} (top). It can be observed that the behavior of the system does not visibly change until a packet loss rate of about $0.3$. This means that the proposed system is quite robust against packet loss. Only for packet loss rates greater than about $0.3$ the system behavior degrades. 

\subsubsection{Limited Energy Harvesting}

Another interesting question concerns the possible change in system behavior when energy harvesting is restricted. Considering the example of solar panels, this is the case, for example, with bad weather conditions. Remember that the proposed model is able to simulate bad weather conditions by means of cloud densities greater than zero. We repeated the initial experiment (that is, without considering packet loss) for a range of different cloud densities between $0$ and $1$. Figure~\ref{fig:duty-experiments} (middle) shows the evolution of the obtained mean system activity when moving from low to high cloud densities. As expected, with increasing cloud density the mean system activity decreases. Interestingly, the relation between cloud density and the mean system activity is linear. 

\subsubsection{Adapting to Different Network Sizes}

When changing to differently sized networks (we only considered 120 node networks so far), it is intuitively clear that some parameter values must be adjusted in order to maintain a functional system. Note that when changing the network size, the node density changes. Hence, it is reasonable to assume that for maintaining the shape of the activity peaks, the choice of the transmission power level and the probability of spontaneous activation should be adapted to the new network size. A way of obtaining the new system parameters is described in the following. With $k_{\mbox{\tiny new}}$, ${p^a}_{\mbox{\tiny new}}$ and $t^{\mbox{\tiny new}}$ we refer to the new number of sensor nodes, the probability of spontaneous activation and the transmission power level of the new, differently sized, network. First, in order to obtain the same wake-up rates as in the case of a 120-node network, the following rule can be applied: ${p^a}_{\mbox{\tiny new}} := p^a \cdot k / k_{\mbox{\tiny new}}$, where $p^a$ and $k$ are the parameters from the original network. Note that this rule increases the probability of spontaneous activation of the nodes when the network size is decreased, and vice-versa when the number of nodes increases. Moreover, the average number of nodes' spontaneous activations per time unit is maintained. Next we introduce a rule for adapting the transmission power level. The basic idea is to have a constant average number of sensors being reached by a transmission. Due to the fact that the sensor nodes form a random topology, the following reasoning was used. In general, the number of nodes that can be reached by the ideal transmission of a sensor can be estimated as follows: $\pi \cdot t \cdot 2 \cdot k/S$, where $k$ is the number of sensor nodes and $S$ is the space in which the sensor nodes reside. In our case it holds that $S = 1^2 = 1$. Therefore, this term reduces to $\pi \cdot {t} \cdot 2 \cdot k$. As $t$ is known for the case of 120-node networks, an adjusted transmission power level can be calculated for networks of different sizes as follows:
\begin{equation}
\label{transmission-power-scaling}
 t^{\mbox{\tiny new}} = \sqrt{\frac{t \cdot 2 \cdot k}{k_{\mbox{\tiny new}}}} \enspace,
\end{equation}
where $k_{\mbox{\tiny new}}$ is the size, respectively $t^{\mbox{\tiny new}}$ is the transmission power level, of the new network. However, note that the transmission power level is not directly modifiable as an algorithm parameter. The only parameters of our algorithm for what concerns to the transmission power level are $\pmin$ and $\pmax$. These values are used as the boundaries of the region for the ideal transmission power levels. Therefore, our scaling method consists in using Eq.~\ref{transmission-power-scaling} for obtaining $\pmin^{\mbox{\tiny new}}$ and $\pmax^{\mbox{\tiny new}}$ for delimiting the value of the new networks' ideal transmission power level.

\begin{figure}[!t]
\centering
  \vspace{-0.5cm}
  \includegraphics[width=5cm,angle=-90]{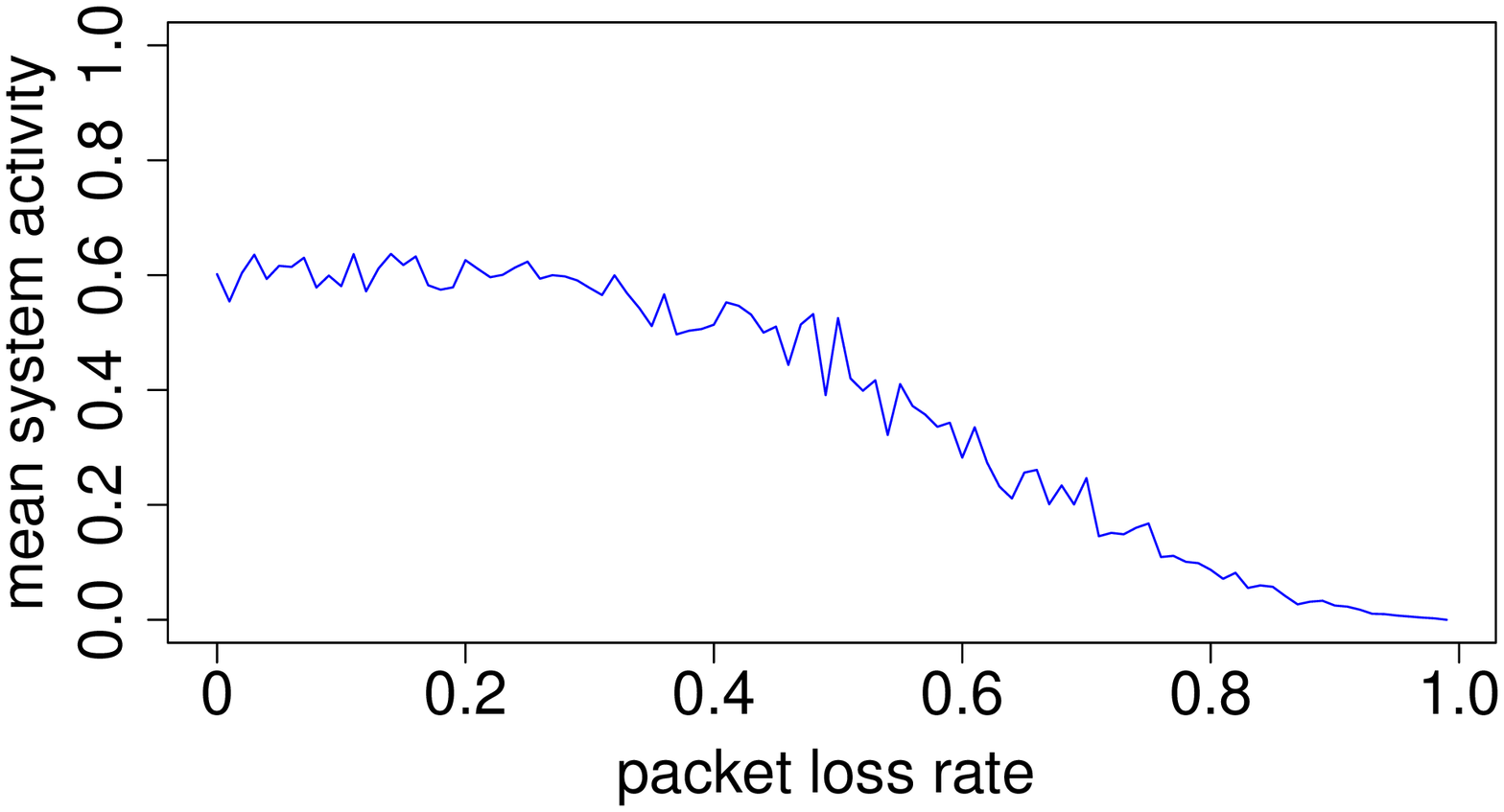} \\
  \vspace{-0.5cm}
  \includegraphics[width=5cm,angle=-90]{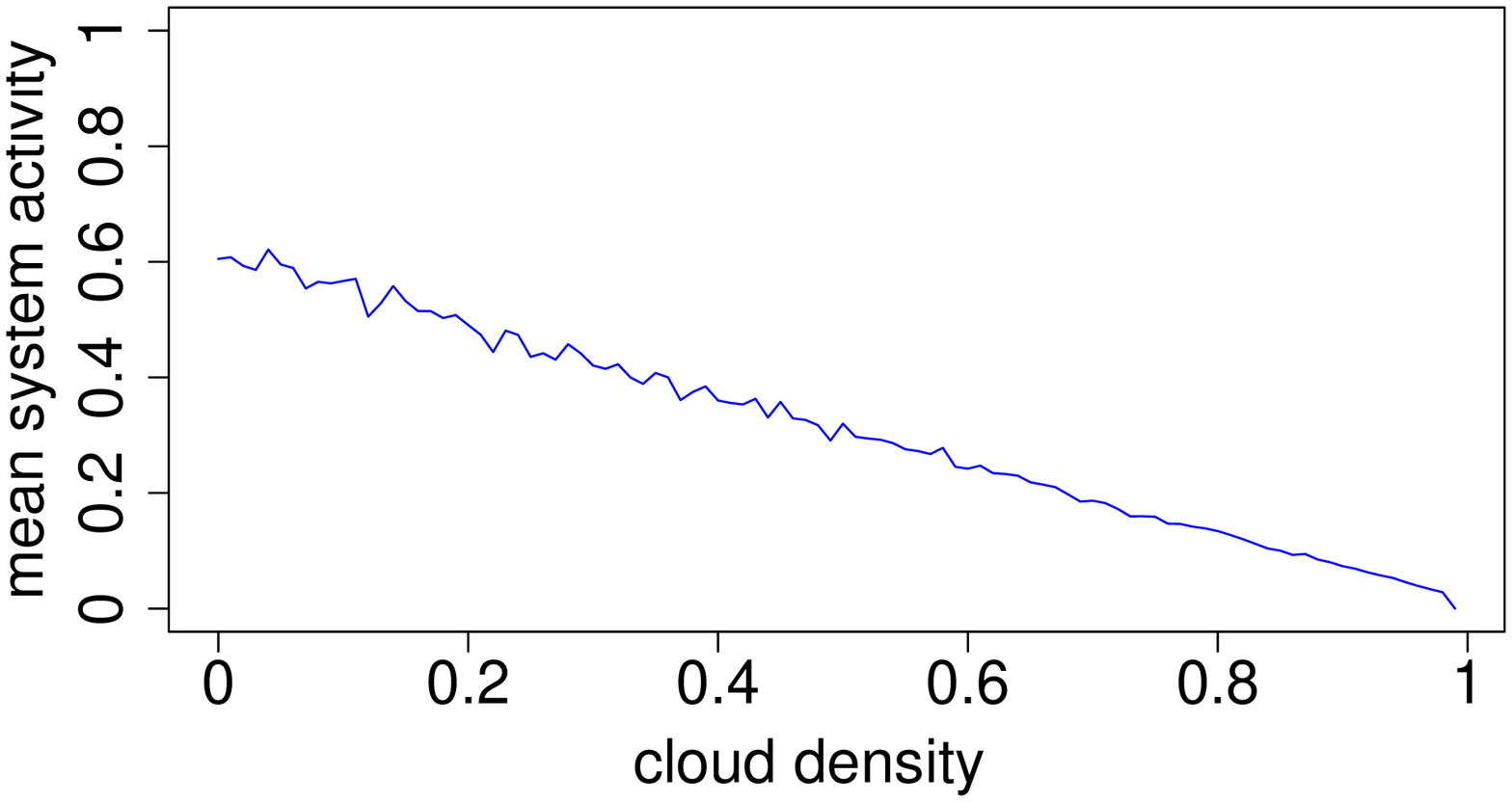} \\
  \vspace{-0.5cm}
  \includegraphics[width=5cm,angle=-90]{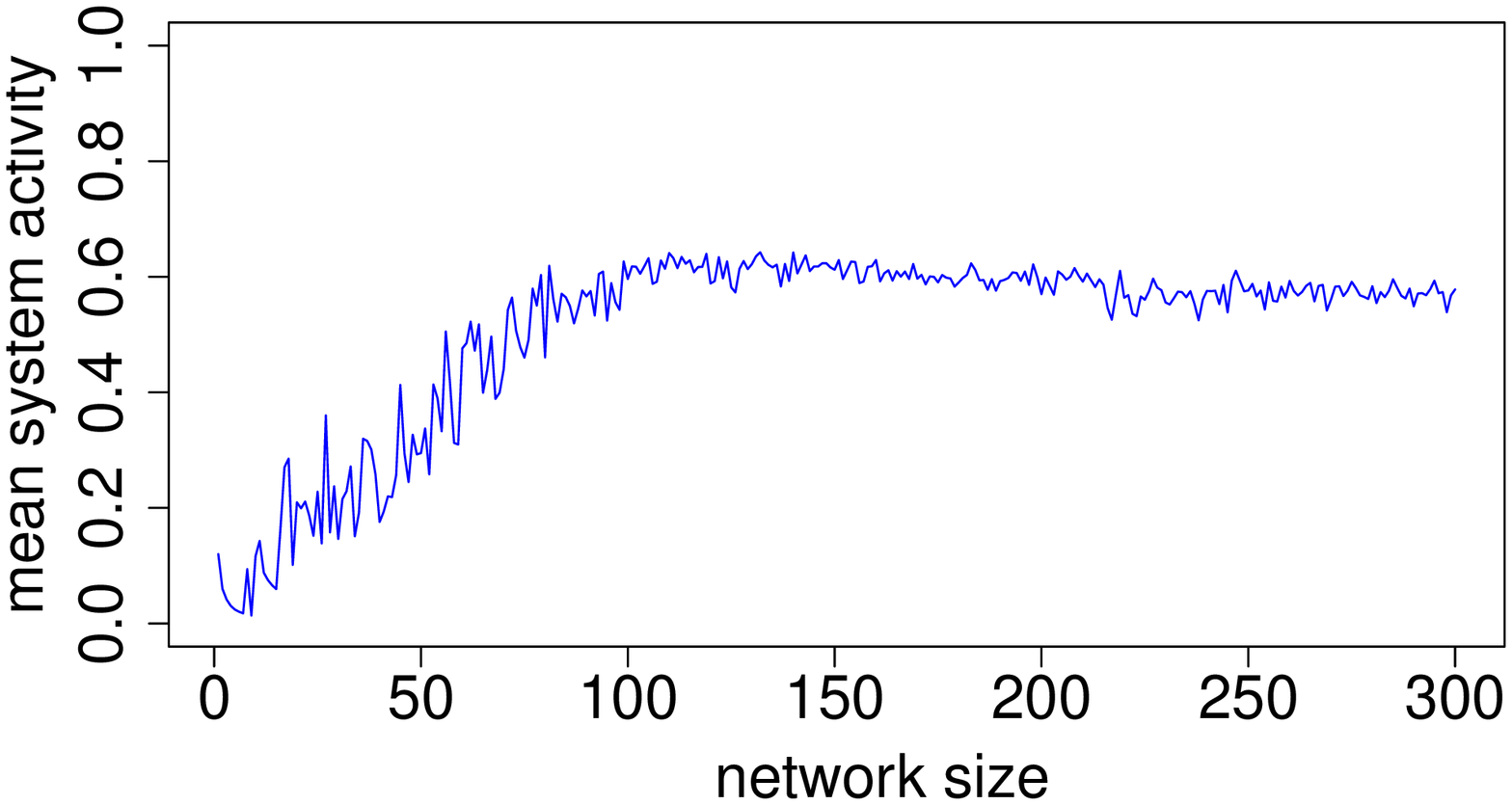}
\caption{Behavior of the duty-cycling mechanism under varying conditions. Top figure: different packet loss rates. Middle figure: different cloud densities. Bottom figure: different network sizes.}
\label{fig:duty-experiments}
\end{figure}

With this scaling methodology we repeated the initial experiment (that is, without packet loss) for a range of different network sizes $k \in [0,300]$. The graphic in Figure~\ref{fig:duty-experiments} (bottom) shows the evolution of the resulting mean system activity. Ideally we would have expected a more or less straight line at about $0.6$. This would have meant that the introduced parameter scaling method leads to a system that behaves equally for all network sizes. However, as can be seen, the scaling method only works well for networks with more than 100 nodes. For smaller networks, the mean system activity decreases. However, this can be explained by the decreasing connectivity and communication ability.

\section{Conclusions and future work}
    \label{sec:conclusions}

In this paper we introduced a protocol for self-synchronized duty-cycling in wireless sensor networks with energy harvesting capabilities. This protocol, inspired by real ant colonies, has been implemented in the generic algorithm library Wiselib. Moreover, experiments have been performed with the network simulator Shawn configured to simulate iSense hardware. The proposed technique adapts to changing energy conditions in a self-organized way. Moreover, it is very robust for what concerns packet loss and limitations of energy harvesting. In the future we plan to combine this protocol with real user applications such as monitoring or tracking. Moreover, we plan to verify the experiments on real hardware. Finally, as our technique does not depend on nodes being static, the results obtained in this paper can be reproduced for mobile sensor networks.

\section*{Acknowledgment}

This work was supported by the EU project FRONTS (FP7-ICT-2007-1). In addition, Christian Blum acknowledges support from the Spanish \textit{Ram{\'o}n y Cajal} program, and Hugo Hern{\'a}ndez acknowledges support from a Catalan \textit{FI} grant.


\end{document}